\documentclass{article} 
\usepackage{iclr2027_conference,times}

\usepackage{amsmath,amsfonts,bm}

\def\eqref#1{equation~\ref{#1}}

\def\1{\bm{1}}

\DeclareMathAlphabet{\mathsfit}{\encodingdefault}{\sfdefault}{m}{sl}
\SetMathAlphabet{\mathsfit}{bold}{\encodingdefault}{\sfdefault}{bx}{n}

\usepackage{hyperref}
\usepackage{url}
\usepackage{graphicx}
\usepackage{amsmath,amssymb}
\usepackage{algorithm}
\usepackage{algpseudocode}
\usepackage{capt-of}

\usepackage{marvosym}  

\usepackage{booktabs}
\usepackage{multirow}
\newcommand{\tablestyle}{%
  \normalsize
  \setlength{\tabcolsep}{3pt}%
  \renewcommand{\arraystretch}{1.0}%
  \setlength{\heavyrulewidth}{0.8pt}%
  \setlength{\lightrulewidth}{0.4pt}%
  \setlength{\cmidrulewidth}{0.4pt}%
  \setlength{\aboverulesep}{0.4ex}%
  \setlength{\belowrulesep}{0.4ex}%
}

\title{UltraWorld: Learning Interactive Ultrasound World Models from Untracked Clinical Videos with Acoustic Sampling Map}

\author{
Keke Yang\textsuperscript{*1}, 
Erqi Wang\textsuperscript{*1}, 
Sainan Guan \textsuperscript{2} \& 
Hongliang Ren \textsuperscript{1 \Letter} \\
\textsuperscript{1} The Chinese University of Hong Kong \\
\textsuperscript{2} The Eighth Affiliated Hospital, Sun Yat-sen University
\\
}

\iclrfinalcopy 
\begin{document}

\maketitle
\lhead{\textit{Preprint. Under review.}}  

\begingroup
\renewcommand{\thefootnote}{}
\footnotetext{
\textsuperscript{*}Equal contribution;
\Letter~Corresponding author.
}
\endgroup

\begin{abstract}

World models can enable autonomous ultrasound scanning by predicting the outcomes of probe motions from local observations. Learning this action--observation relationship typically relies on synchronized video--pose pairs, which are costly to collect at scale and largely unavailable in routine clinical recordings. Reliable action following further requires modeling ultrasound's cross-sectional sampling geometry.
We present UltraWorld, a self-distillation recipe that transfers priors from clinical ultrasound videos into interactive world models without real action annotations. Starting from clinical videos, we adapt a video foundation model into an ultrasound generator conditioned on reference images and anatomical masks. Anatomical masks sampled along programmable trajectories through 3D anatomy provide spatial guidance for synthesizing action--video pairs. We then use these synthetic pairs to self-distill the generator into a world model that predicts future observations from local observations and actions, without requiring anatomical masks or other 3D assets at inference time. To further improve action following, we introduce the Acoustic Sampling Map (AsMap), which represents probe poses and imaging settings as pixel-wise 3D sampling positions, beam directions, and depths.
Experiments demonstrate improved prediction fidelity and action following. Across nine simulated closed-loop local planning episodes, UltraWorld reduces the mean final distance to the goal and orientation error by 29\% and 38\%, respectively, compared with visual servoing.
Project Page: \url{https://ultraworld-project.github.io/}.
\end{abstract}

\section{Introduction} 

World models have shown promise for autonomous ultrasound scanning, supporting probe guidance \citep{yue2025echoworld} and policy learning from predicted observations \citep{fan2026actionconditioned}. 
Predicting the outcomes of candidate actions from local observations supports planning and policy evaluation or improvement through simulated interactions \citep{hafner2020dreamer,huang2026vid2world,guo2026ctrl}. This is particularly useful in ultrasound, where sonographers face the demanding task of relating 2D sections to 3D anatomy to guide probe motion \citep{jiang2025towards}.
Two challenges limit this capability. First, acquiring synchronized ultrasound video--pose data is costly, limiting data scale and clinical diversity. Second, scanning operations alter which anatomical cross-sections are imaged and how they are sampled, making reliable action-controlled prediction difficult in ultrasound.

\begin{figure}[t]
    \centering
    \includegraphics[width=\linewidth]{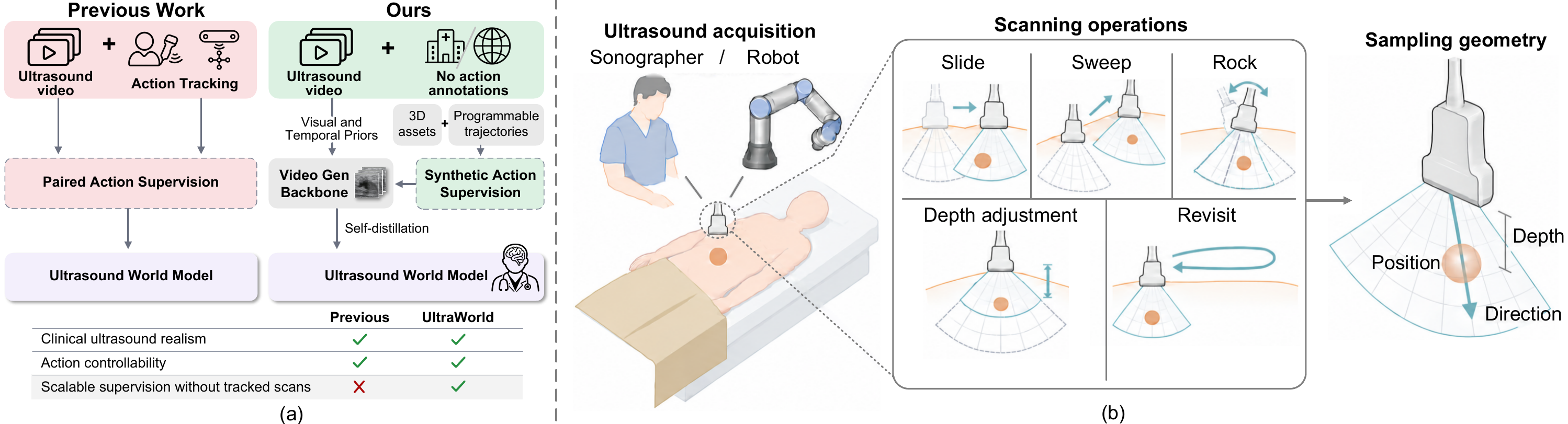}
    \caption{
    \textbf{Learning ultrasound world models from untracked clinical videos.}
    \textbf{(a)} Clinical videos provide appearance priors, while 3D anatomy provides motion supervision. Together, they yield synthetic action--video pairs for self-distillation.
    \textbf{(b)} Complex scanning operations change which anatomical cross-sections are imaged and how they are sampled. AsMap encodes the resulting sampling positions, beam directions, and depths to support reliable action control.
    }
    \label{fig:uswm_concept}
\end{figure}

Recent world models transfer pretrained video generators for data efficiency but still require observation--action pairs to learn target-domain control \citep{huang2026vid2world}.
In ultrasound, collecting these pairs requires expert operators, calibrated tracking, synchronized recording, and participant recruitment, making large-scale acquisition labor-intensive. Existing clinical videos, meanwhile, usually lack probe-motion records.   
Simulation offers controlled interactions, but full-path ray tracing may omit tissue scattering and attenuation \citep{duelmer2025ultraray}, while learning-based simulation may require specially acquired and registered CT--ultrasound pairs \citep{ao2025sonogym}.
This motivates a more scalable route to learning reliable action responses from untracked clinical videos.

Our key insight is that visual and temporal priors learned from clinical videos can be transferred into an interactive world model using independently constructed action supervision (Figure~\ref{fig:uswm_concept}(a)). 
Clinical videos capture anatomical appearance and temporal changes \citep{mishra2026self}; their plane selection and scanning sequences also implicitly reflect clinical judgment and expertise. 
We can learn these priors in a controllable video generator, then construct action supervision using 3D anatomical assets and programmable scanning trajectories. Self-distillation provides a route for this transfer, using generated outputs as supervision to learn new conditioning capabilities \citep{cai2025diffusion} or representations \citep{bahmani2026lyra}. Using generated action--video pairs, we can transfer this prior to prediction conditioned on local observations and scanning operations, without tracked clinical recordings.

Learning from this supervision also requires relating local 2D observations to probe motion in 3D: scanning operations change both the anatomical cross-section and its sampling geometry (Figure~\ref{fig:uswm_concept}(b)). Pose-only inputs omit imaging-scale changes, while camera encodings describe viewing rays or projective relationships \citep{li2025prope}. Ultrasound pixels instead correspond to depth-localized samples along acoustic beams \citep{szabo2014ultrasound}. Reliable action conditioning therefore needs to make these sampling locations explicit under probe motion and imaging adjustments.

We present UltraWorld, a recipe for learning interactive ultrasound world models from untracked clinical videos. We adapt a video foundation model into an ultrasound generator conditioned on a reference image and anatomical mask sequences. For action supervision, we generate action--video pairs using mask sequences sampled from 3D assets along programmable trajectories. 
Through self-distillation, we transfer the learned ultrasound priors into an observation- and action-conditioned world model initialized from the generator's weights, requiring neither masks nor 3D assets at inference. To link 3D probe motion to changes in 2D observations, we introduce the Acoustic Sampling Map (AsMap), encoding probe poses and imaging settings as pixel-wise 3D sampling positions, beam directions, and depths. Experiments show improved prediction fidelity and action following. We further evaluate consistency from unseen clinical frames and explore closed-loop local planning for robotic ultrasound scanning in simulation.

Our main contributions are as follows:
(1) A \textbf{self-distillation recipe for learning ultrasound world models from untracked clinical videos}, combining clinical video priors with programmable supervision from 3D anatomy to learn action-conditioned prediction without real video--pose pairs.
(2) The \textbf{Acoustic Sampling Map (AsMap)}, which grounds probe motion and imaging adjustments in pixel-wise 3D sampling positions, beam directions, and depths to support reliable action control in cross-sectional ultrasound imaging.
(3) \textbf{Experimental validation and a robotic planning application}, demonstrating improved prediction fidelity and action following, evaluating consistency from unseen clinical frames, and exploring closed-loop local planning for robotic ultrasound scanning in simulation.

\section{Related works}
\label{sec:related_work}

\paragraph{Robotic Ultrasound.}
Robotic ultrasound guidance has been explored through visual
servoing~\citep{abolmaesumi2002image,chatelain2017confidence,ma2025guiding},
reinforcement learning~\citep{li2021autonomous}, and demonstration-based
reward learning~\citep{jiang2024intelligent}.
Integrated systems further combine learned navigation, probe-orientation
optimization, and force feedback for autonomous thyroid
scanning~\citep{su2024fully}.
More recently, ultrasound world models have demonstrated the value of
predicting acquisition outcomes for improving probe guidance and supporting
policy learning.
Cardiac Copilot and EchoWorld learn cardiac world models for
echocardiography probe guidance~\citep{jiang2024cardiac,yue2025echoworld},
while \citet{fan2026actionconditioned} use a diffusion world model to provide rewards
for goal-conditioned policy training and demonstrate closed-loop goal-plane
navigation.
Despite these benefits, both world-model approaches rely on synchronized
ultrasound--pose recordings to learn acquisition dynamics, leaving routine
untracked clinical videos underexploited.

\paragraph{Ultrasound Synthesis.}
Ultrasound synthesis encompasses anatomical simulation from CT/MR-derived anatomy and learned generation \citep{kutter2009visualization,salehi2015patient}.
SonoGym~\citep{ao2025sonogym} supports physics-based and generative ultrasound
simulation from CT-derived anatomy for robotic learning.
Echo from Noise~\citep{stojanovski2023echo} uses semantic-conditioned diffusion
to generate training images for cardiac segmentation, while
BUSGen~\citep{yu2026busgen} learns large-scale breast ultrasound priors to synthesize task-specific data for downstream analysis.
Video diffusion models generate cardiac cycles from semantic
maps~\citep{nguyen2023echovideo} or arbitrary single
frames~\citep{zhang2026echovdiff}.
These studies demonstrate the value of anatomical control and synthetic data,
but visual realism alone does not establish faithful responses to prescribed
scanning actions. Connecting clinical video priors with acquisition control
requires consistency between the commanded trajectory and the evolving anatomy.

\paragraph{Geometric Conditioning.}
Geometric conditioning encodes pose and imaging geometry within visual models.
Ultrasound world models encode probe motion using learned embeddings or
multi-scale trigonometric features~\citep{yue2025echoworld,fan2026actionconditioned},
leaving pixel-wise sampling relationships implicit.
In camera-controlled video generation, CameraCtrl II~\citep{he2025cameractrlii} uses pixel-aligned Pl\"ucker embeddings with input-level feature addition, 
while VD3D~\citep{bahmani2025vd3d} injects Pl\"ucker camera embeddings into video diffusion transformers through a ControlNet-like conditioning mechanism.
GTA~\citep{miyato2024gta} incorporates relative geometric transformations into
attention, while PRoPE~\citep{li2025prope} models relative camera-frustum
relationships through both intrinsics and extrinsics.
These camera-centric representations encode optical viewing rays or projective
relationships. Adapting such conditioning to ultrasound requires representing depth-resolved acoustic sampling under probe motion and imaging configuration.

\section{Preliminaries}
\label{sec:preliminaries}

\paragraph{Controllable video generation.}
Following controllable video models~\citep{jiang2025vace}, we denote a
reference condition by $\mathbf{r}$ and temporally aligned controls by
$\mathbf{C}_{1:T}=\{\mathbf{c}_t\}_{t=1}^{T}$. A conditional video generator
models
$p_{\theta}(\mathbf{X}_{1:T}\mid\mathbf{r},\mathbf{C}_{1:T})$,
where $\mathbf{X}_{1:T}=\{\mathbf{x}_t\}_{t=1}^{T}$ denotes the generated video.

\paragraph{Ultrasound acquisition parameters.}
Clinical ultrasound acquisition is determined jointly by probe pose and imaging
configuration. We denote the acquisition state at time $t$ as $\boldsymbol{\xi}_t = (\mathbf{T}_t, \boldsymbol{\kappa}_t)$,
where $\mathbf{T}_t\in SE(3)$ transforms points from the probe coordinate
system to the patient coordinate system, and $\boldsymbol{\kappa}_t$ denotes
acquisition parameters such as scan geometry, physical sampling scale, imaging
depth or field of view, and beam steering.

An ultrasound acquisition maps a physical point $\mathbf{p}$ to its image
coordinate as
\begin{equation}
    (u,v)
    =
    \pi_{\mathrm{US}}
    \left(
        \mathbf{T}_t\mathbf{p};
        \boldsymbol{\kappa}_t
    \right),
    \label{eq:us_projection}
\end{equation}
where $\pi_{\mathrm{US}}$ denotes the acquisition-dependent mapping from
spatial locations on the imaging plane to ultrasound image coordinates.

\paragraph{World models.}
A visual world model predicts the visual consequences of actions from an
observed context~\citep{zhu2025irasim,huang2026vid2world}. Given
$\mathbf{X}^{c}=\{\mathbf{x}_1,\ldots,\mathbf{x}_K\}$ and future actions
$\mathbf{A}^{f}=\{\mathbf{a}_K,\ldots,\mathbf{a}_{K+H-1}\}$, it predicts
$\widehat{\mathbf{X}}^{f}
=\mathcal{W}_{\theta}(\mathbf{X}^{c},\mathbf{A}^{f})$, where
$\widehat{\mathbf{X}}^{f}
=\{\hat{\mathbf{x}}_{K+1},\ldots,\hat{\mathbf{x}}_{K+H}\}$.

\section{Method}

\begin{figure}[t]
    \centering
    \includegraphics[width=\linewidth]{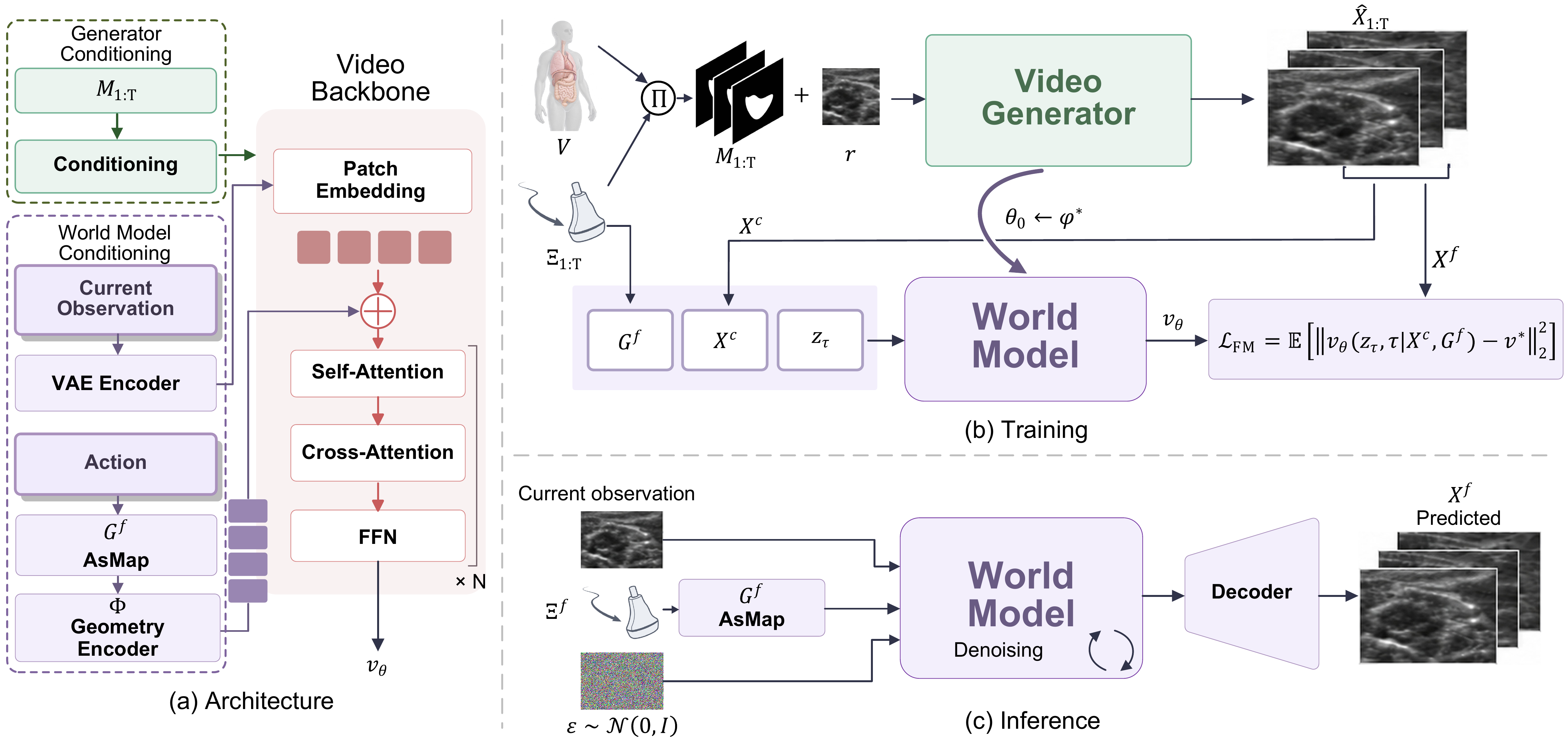}
    \caption{
    \textbf{Overall framework of UltraWorld.}
    (a) A shared video backbone supports mask-conditioned generation and action-conditioned world modeling through AsMap.
    (b) 3D anatomy and programmable trajectories provide synthetic action--video supervision, transferring the learned video prior to the world model through self-distillation and flow matching.
    (c) At inference, the world model predicts future ultrasound video from the current observation and prescribed acquisition trajectory.
    }
    \label{fig:pipeline}
\end{figure}

In this section, we present \textbf{UltraWorld}, a recipe for learning interactive
ultrasound world models from action-free clinical videos.
Our key idea is to transfer the visual and temporal priors learned from clinical
videos into action-conditioned dynamics through anatomy-guided interaction
supervision. We further introduce a pixel-aligned representation of ultrasound
acquisition to explicitly encode the spatial effects of diverse clinical scanning
operations. The overall approach is illustrated in Fig.~\ref{fig:pipeline}.

\subsection{From Action-Free Ultrasound Videos to Interactive Trajectories}
\label{sec:interactive_trajectory}

We reformulate world-model learning as transferring priors from action-free clinical ultrasound videos to trajectories with explicit acquisition supervision. We first learn a controllable video prior, then construct the missing action supervision independently from anatomy-guided scanning trajectories.

\paragraph{Learning ultrasound priors from action-free videos.}
Given an action-free ultrasound sequence
$\mathbf{X}_{1:T}=\{\mathbf{x}_t\}_{t=1}^{T}$,
we associate it with a temporally aligned semantic anatomical condition
$\mathbf{M}_{1:T}=\{\mathbf{m}_t\}_{t=1}^{T}$.
Instantiating the structured video control
$\mathbf{C}_{1:T}$ with $\mathbf{M}_{1:T}$, we learn a controllable ultrasound
video generator
\begin{equation}
    p_{\phi}
    \left(
        \mathbf{X}_{1:T}
        \mid
        \mathbf{r},
        \mathbf{M}_{1:T}
    \right),
    \label{eq:mask_generator}
\end{equation}
where $\mathbf{r}$ is a reference ultrasound image that anchors the appearance
characteristics of the generated sequence.
Crucially, this stage requires no corresponding acquisition states
$\boldsymbol{\Xi}_{1:T}$, allowing ultrasound-specific visual and temporal
priors to be learned directly from action-free clinical videos.

\paragraph{Anatomy-guided ultrasound scanning trajectories.}
The learned video prior captures realistic ultrasound appearance and temporal
evolution, but action-free videos do not specify how observations should change
under prescribed acquisition states. We obtain this missing interaction
supervision from 3D anatomical volumes.
Let $\mathcal{V}$ denote a semantic anatomical volume. We construct a scanning
trajectory as a sequence of acquisition states
\begin{equation}
    \boldsymbol{\Xi}_{1:T}
    =
    \left\{
        \boldsymbol{\xi}_t
    \right\}_{t=1}^{T},
    \qquad
    \boldsymbol{\xi}_t
    =
    \left(
        \mathbf{T}_t,
        \boldsymbol{\kappa}_t
    \right),
    \label{eq:scanning_trajectory}
\end{equation}
where $\mathbf{T}_t$ denotes the transducer pose and
$\boldsymbol{\kappa}_t$ specifies the ultrasound acquisition parameters.
We programmatically sample trajectories covering common clinical scanning
operations, including sweeps, slide, rock, revisits, and changes in imaging depth
or field of view.

For each acquisition state, the corresponding anatomical observation is obtained
by resampling the 3D volume under the specified acquisition configuration,
\begin{equation}
    \mathbf{m}_t
    =
    \Pi
    \left(
        \mathcal{V};
        \boldsymbol{\xi}_t
    \right),
    \label{eq:volume_projection}
\end{equation}
where $\Pi(\cdot)$ denotes the volumetric sampling operator.
Applying this operation along $\boldsymbol{\Xi}_{1:T}$ yields an anatomically
consistent mask trajectory $\mathbf{M}_{1:T}$ whose acquisition states are known
by construction. Consistent with prior ultrasound simulation, we use MR only to provide 3D anatomical geometry for \(M_{1:T}\), while ultrasound appearance is inherited from the clinical-ultrasound prior \(p_\phi\) \citep{kutter2009visualization}.

\paragraph{Interactive trajectory synthesis.}
We then use the anatomy-guided mask trajectory as the structured control of the
learned ultrasound video prior:
\begin{equation}
    \hat{\mathbf{X}}_{1:T}
    \sim
    p_{\phi}
    \left(
        \mathbf{X}_{1:T}
        \mid
        \mathbf{r},
        \mathbf{M}_{1:T}
    \right).
    \label{eq:interactive_generation}
\end{equation}
Because $\mathbf{M}_{1:T}$ is generated from the prescribed acquisition
trajectory $\boldsymbol{\Xi}_{1:T}$, the synthesized ultrasound sequence is
paired by construction with its acquisition states. Repeating this process
produces an interaction dataset
\begin{equation}
    \hat{\mathcal{D}}_{\mathrm{traj}}
    =
    \left\{
        \left(
            \hat{\mathbf{X}}_{1:T},
            \boldsymbol{\Xi}_{1:T}
        \right)
    \right\},
    \label{eq:interaction_dataset}
\end{equation}
thereby converting the visual and temporal priors learned from action-free
clinical videos into scalable action-labeled supervision for ultrasound
world-model learning.

\subsection{Acoustic Sampling Map}
\label{sec:asmap}

\begin{figure}[t]
    \centering
    \includegraphics[width=0.8\linewidth]{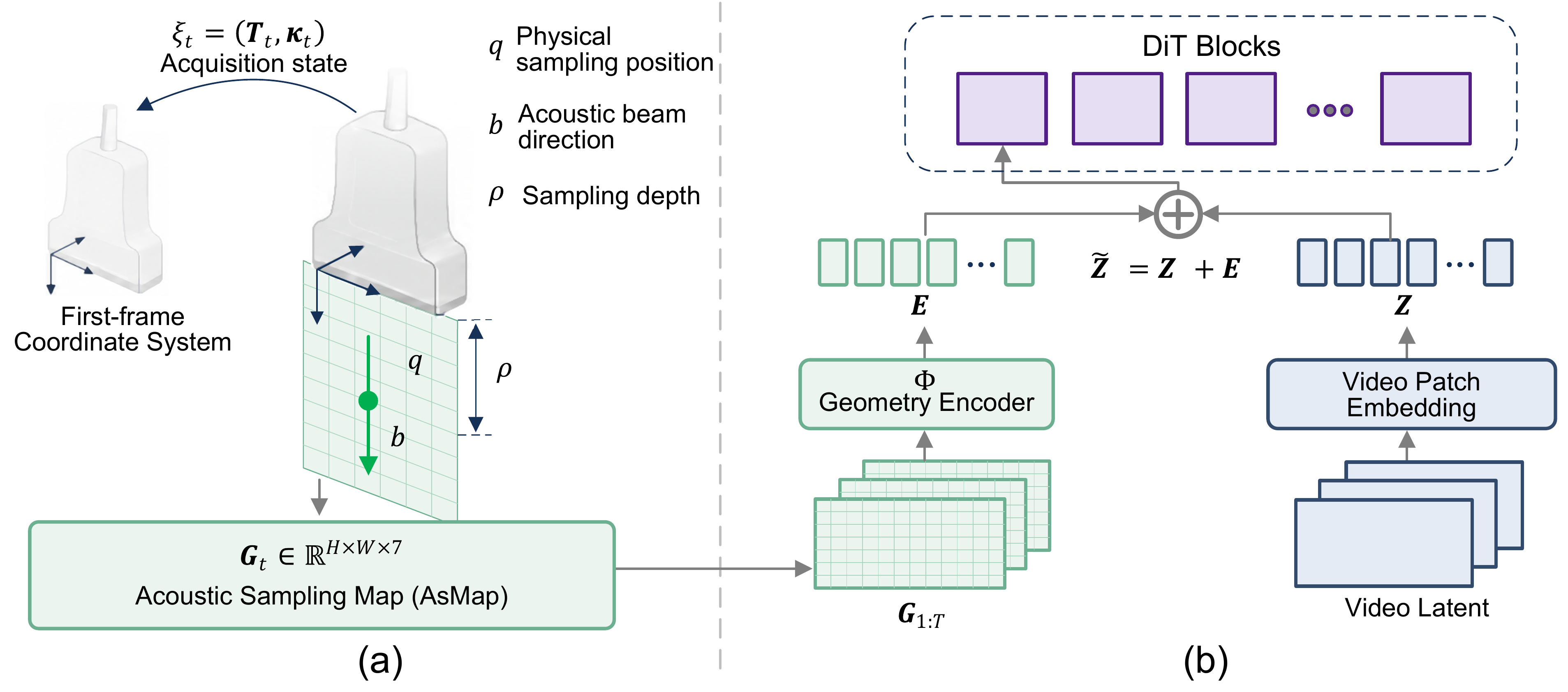}
    \caption{
    \textbf{Acoustic Sampling Map and geometry conditioning.}
    (a) Probe pose and imaging parameters determine each pixel's physical sampling
    position $\mathbf{q}$, beam direction $\mathbf{b}$, and sampling depth $\rho$.
    Positions and directions are expressed in the first-frame probe coordinate
    system, forming a seven-channel AsMap.
    (b) A geometry encoder $\Phi$ aligns the AsMap sequence with the
    video tokens, providing pixel-aligned conditioning before the first
    transformer block.}
    \label{fig:asmap}
\end{figure}

Clinical ultrasound acquisition depends jointly on probe motion and imaging
configuration. For example, a sonographer may reduce the imaging depth to
enlarge a lesion within the field of view while keeping the probe pose nearly
unchanged. A 6-DoF pose alone cannot distinguish these acquisitions, although
they correspond to different physical sampling geometries and therefore
different image observations. To explicitly represent the spatial sampling
induced by both probe motion and acquisition parameters, we introduce the
\emph{Acoustic Sampling Map} (AsMap), a dense, pixel-aligned representation
of ultrasound acquisition geometry (Fig.~\ref{fig:asmap}).

\paragraph{Acoustic sampling geometry.}
For each image location $(u,v)$, we define its probe-local sampling geometry as
\begin{equation}
\mathcal{S}_{\mathrm{US}}(u,v;\kappa_t)
=
\left(
\mathbf{q}_{t,u,v},
\mathbf{b}_{t,u,v},
\rho_{t,u,v}
\right),
\label{eq:asmap_sampling}
\end{equation}
where $\mathbf{q}_{t,u,v}\in\mathbb{R}^3$ is the physical sampling position,
$\mathbf{b}_{t,u,v}\in\mathbb{R}^3$ is the unit acoustic beam direction, and
$\rho_{t,u,v}$ is the sampling depth.

To describe a trajectory consistently, we express positions and directions in
the probe coordinate system of the first frame. With
$T_t=(R_t,\mathbf{t}_t)$ denoting the probe-to-patient transformation,
\begin{equation}
\begin{bmatrix}
\mathbf{q}^{(1)}_{t,u,v}\\1
\end{bmatrix}
=
T_1^{-1}T_t
\begin{bmatrix}
\mathbf{q}_{t,u,v}\\1
\end{bmatrix},
\qquad
\mathbf{b}^{(1)}_{t,u,v}
=
R_1^\top R_t\mathbf{b}_{t,u,v}.
\label{eq:asmap_relative}
\end{equation}
The AsMap at time $t$ is then
\begin{equation}
\mathbf{G}_t(u,v)
=
\left[
\mathbf{q}^{(1)}_{t,u,v},
\mathbf{b}^{(1)}_{t,u,v},
\rho_{t,u,v}
\right]
\in\mathbb{R}^{7},
\qquad
\mathbf{G}_t\in\mathbb{R}^{H\times W\times7}.
\label{eq:asmap}
\end{equation}
Unlike pose-only conditioning, AsMap explicitly changes with the physical
sampling configuration even when the probe pose is fixed. The first-frame
formulation also removes dependence on the arbitrary global patient coordinate
frame; additional properties are discussed in
Appendix~\ref{app:asmap_properties}.

\paragraph{Encoding and injection.}
A lightweight geometry encoder $\Phi$ converts the AsMap sequence into
features aligned with the video tokens,
\begin{equation}
\mathbf{E}=\Phi(\mathbf{G}_{1:T}),
\qquad
\widetilde{\mathbf{Z}}=\mathbf{Z}+\mathbf{E},
\label{eq:asmap_injection}
\end{equation}
where $\mathbf{Z}$ denotes the patch-embedded video features. Following the
input-level conditioning strategy of CameraCtrl~II~\citep{he2025cameractrlii},
the AsMap features are added once before the first transformer block, providing
frame- and pixel-aligned acquisition conditioning without modifying the
internal transformer architecture.

\subsection{Self-Distillation for Ultrasound World Modeling}
\label{sec:self_distillation}

\begin{figure}[!t]
    \centering
    \includegraphics[width=\linewidth]{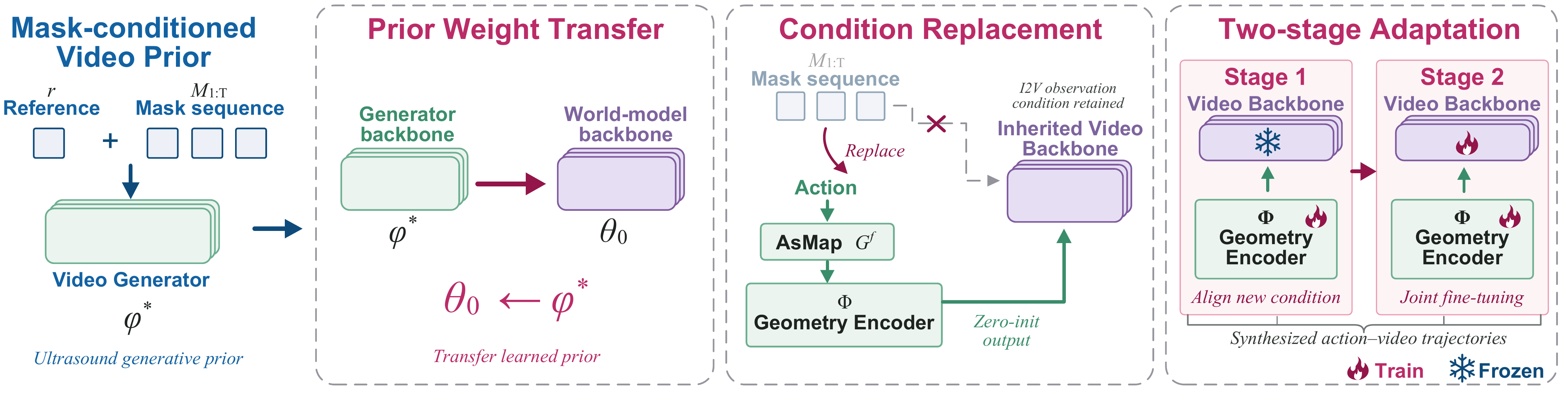}
    \caption{
    \textbf{Self-distillation for ultrasound world modeling.}
    The transformation consists of three steps:
    (1) \textbf{Prior weight transfer}: initialize the world model
    from the learned video generator;
    (2) \textbf{Condition replacement}: replace anatomical-mask
    conditioning with AsMap conditioning while retaining the observation input;
    (3) \textbf{Two-stage adaptation}: train the geometry encoder
    with a frozen video backbone, then jointly fine-tune both.
    Both stages use synthesized action--video trajectories.
    }
    \label{fig:self_distillation}
\end{figure}

The synthesized trajectories transfer geometry-controlled video generation into action-conditioned prediction of how ultrasound appearance changes with scanning operations.
Following generative self-distillation~\citep{cai2025diffusion,bahmani2026lyra},
we transform the mask-conditioned ultrasound generator into an
action-conditioned world model by replacing mask conditioning with AsMap while retaining the observation input (Fig.~\ref{fig:self_distillation}). We initialize the world-model backbone from the teacher, $\boldsymbol{\theta}_{0}\leftarrow\boldsymbol{\phi}^{*}$, and train it on the synthesized trajectories to model
$p_{\theta}(\mathbf{X}^{f}\mid\mathbf{X}^{c},\mathbf{G}^{f})$. For optimization, we first train the AsMap encoder $\Phi$ with the video backbone frozen and then jointly fine-tune both. Detailed training procedures are provided in Appendix~\ref{app:implementation}.

\section{Experiments}
\label{sec:experiments}

\subsection{Experimental Setup}
\label{sec:exp_setup}

\paragraph{Datasets.}
We use BUV~\citep{lin2022buv} and CAMUS~\citep{leclerc2019camus}
to learn the mask-conditioned ultrasound video prior, and
I-SPY1~\citep{newitt2016ispy1,hylton2016ispy1} to construct
anatomy-guided trajectories from volumetric breast MRI. CAMUS is used only for prior learning.
\textbf{Training details.}
We initialize the ultrasound video generator from
Wan2.1-FunControl-1.3B-V1.1~\citep{wan2025} and train the generator
and world model at $512\times512$ resolution.

\paragraph{Baselines.}
We compare AsMap with \textbf{Fourier Encoding}~\citep{fan2026actionconditioned},
\textbf{MLP Encoding} inspired by EchoWorld~\citep{yue2025echoworld}, and
\textbf{PRoPE}~\citep{li2025prope}. All methods receive the same acquisition
state $(\mathbf{T}_t,\boldsymbol{\kappa}_t)$.

\subsection{Mask-Conditioned Ultrasound Video Generator}
\label{sec:mask_video}

\begin{figure}[h]
\begin{center}
\includegraphics[width=0.88\linewidth]{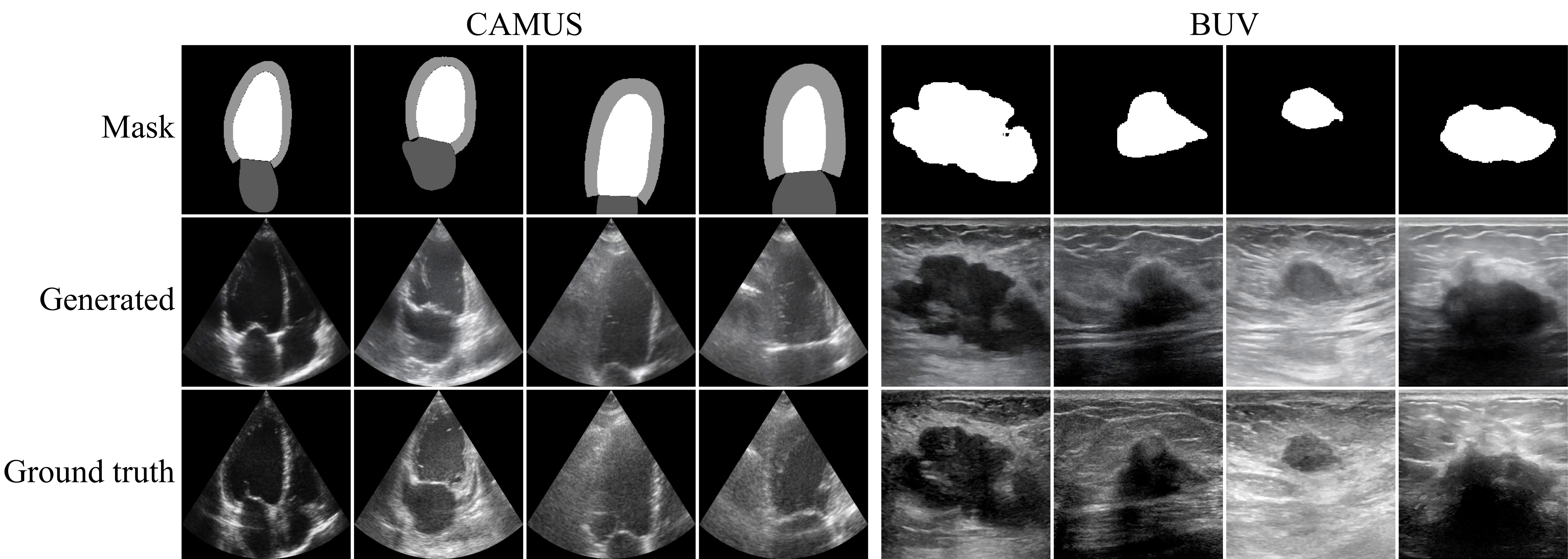}
\end{center}
\caption{
\textbf{Mask-conditioned ultrasound video generation.}
Generated sequences follow the prescribed anatomy while preserving realistic
ultrasound appearance.
}
\label{fig:mask_generator}
\end{figure}

We first verify that the mask-conditioned generator preserves realistic
ultrasound appearance while following the prescribed anatomical structure.
As shown in Fig.~\ref{fig:mask_generator}, the generated sequences remain
temporally coherent and follow the target anatomy across both breast ultrasound
and echocardiography. Quantitative evaluation of generation fidelity and mask
controllability is provided in Appendix~\ref{app:generator_eval}.

\subsection{World Model Quality Analysis}
\label{sec:wm_eval}

We evaluate whether UltraWorld predicts plausible changes in ultrasound texture and anatomical appearance in response to prescribed acquisition trajectories.
Following recent discussions on world-model evaluation
~\citep{yu2026worldmodeleval}, we consider both prediction fidelity and
action-sensitive behavior. We first evaluate on paired synthesized trajectories
with known reference futures and acquisition states.
Counterfactual evaluations under alternative trajectories are provided in
Appendix~\ref{app:results_limitations}.

\paragraph{Metrics.}
Following trajectory-conditioned world models~\citep{zhu2025irasim}, we evaluate
prediction fidelity using PSNR, LPIPS, and latent $\ell_2$ distance, and action
following using $\Delta$-LPIPS~\citep{li2026dworldeval} and \emph{Motion Error}.
The latter measures the discrepancy between predicted and reference
target-anatomy displacements. Let $\boldsymbol{\mu}_t$ and $\hat{\boldsymbol{\mu}}_t$ denote the reference and predicted
target-anatomy centroids at frame $t$:
\begin{equation}
E_{\mathrm{motion}} = \frac{1}{|\mathcal{P}|}\sum_{t\in\mathcal{P}}
\left\| (\hat{\boldsymbol{\mu}}_{t+1}-\hat{\boldsymbol{\mu}}_t) - (\boldsymbol{\mu}_{t+1}-\boldsymbol{\mu}_t) \right\|_2 ,
\label{eql:motion_error}
\end{equation}
where $\mathcal{P}$ denotes the set of valid frame pairs.

\paragraph{Prediction fidelity and action following.}
Table~\ref{tab:wm_quality} compares AsMap with representative global
action encodings and projective geometric conditioning under the same
acquisition states. AsMap achieves the best performance across both
prediction-fidelity and action-following metrics, showing that explicitly
representing pixel-wise ultrasound sampling geometry improves both visual
prediction and responses to prescribed acquisition changes.

\begin{table}[!t]
\caption{
\textbf{Comparison of different action-conditioning representations.}
AsMap consistently improves both prediction fidelity and action following
over global action encodings and projective geometric conditioning.
Motion error is measured in pixels.
}
\label{tab:wm_quality}
\begin{center}
\tablestyle
\begin{tabular*}{\linewidth}{@{\extracolsep{\fill}}lccccc@{}}
\toprule
\multirow{2}{*}{Conditioning} &
\multicolumn{3}{c}{Prediction fidelity} &
\multicolumn{2}{c}{Action following} \\
\cmidrule(lr){2-4}
\cmidrule(l){5-6}
&a
PSNR$\uparrow$ &
LPIPS$\downarrow$ &
Latent $\ell_2\downarrow$ &
$\Delta$-LPIPS$\downarrow$ &
Motion err.$\downarrow$ \\
\midrule
Fourier Enc. \citep{fan2026actionconditioned}
& 22.946 & 0.274 & 0.120 & 0.509 & 4.540 \\
MLP Enc. \citep{yue2025echoworld}
& 23.031 & 0.262 & 0.121 & 0.486 & 4.471 \\
PRoPE \citep{li2025prope}
& 22.634 & 0.275 & 0.127 & 0.504 & 4.457 \\
\midrule
\textbf{AsMap (Ours)}
& \textbf{24.994}
& \textbf{0.236}
& \textbf{0.095}
& \textbf{0.466}
& \textbf{4.155} \\
\bottomrule
\end{tabular*}
\end{center}
\end{table}

\begin{figure}[!t]
\begin{center}
\includegraphics[width=\linewidth]{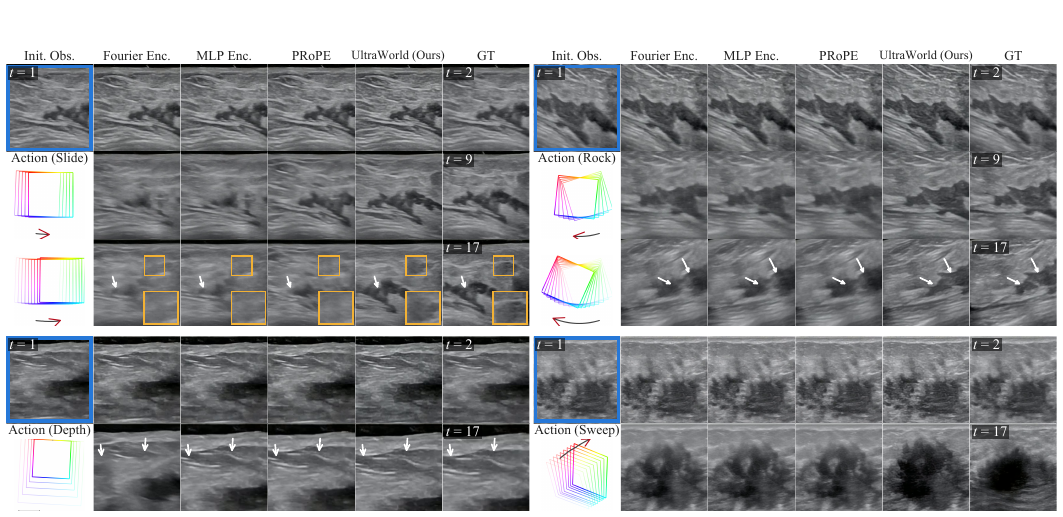}
\end{center}
\caption{
\textbf{Qualitative comparison across scanning operations.}
UltraWorld follows prescribed operations more consistently.
White arrows highlight differences in action following; yellow insets highlight missing or unstable hypoechoic regions and anatomical features in baseline predictions.
}
\label{fig:wm_quality}
\end{figure}

Figure~\ref{fig:wm_quality} compares future ultrasound predictions from the same initial observation under four scanning operations. White arrows highlight where baseline predictions fail to follow the image changes induced by probe motion or imaging adjustments. Yellow insets reveal missing or unstable local features, including hypoechoic regions and surrounding echogenic tissue interfaces. UltraWorld more consistently predicts how these features evolve with the prescribed actions while maintaining plausible texture and local anatomical appearance across future frames. Additional examples are provided in Appendix~\ref{app:appearance_dynamics}.

\paragraph{Ablation studies.}
\label{sec:ablation}
A parameter-matched MLP using the same first-frame-relative coordinates
remains worse than AsMap (Table~\ref{tab:ablation}), indicating that the
gain is not explained by encoder capacity or coordinate convention.
Initialization from Wan2.1 performs comparably, suggesting that
self-distillation does not critically rely on generator-weight inheritance.

\begin{table}[!t]
\caption{
\textbf{Ablation studies on action encoding and initialization.}
``Matched'' approximately matches the AsMap encoder parameter count and uses
the same first-frame-relative coordinates. For each ablation, the remaining
components follow the full model configuration.
}
\label{tab:ablation}
\begin{center}
\tablestyle
\begin{tabular*}{\linewidth}{@{\extracolsep{\fill}}lccccc@{}}
\toprule
\multirow{2}{*}{Variant} &
\multicolumn{3}{c}{Prediction fidelity} &
\multicolumn{2}{c}{Action following} \\
\cmidrule(lr){2-4}
\cmidrule(l){5-6}
&
PSNR$\uparrow$ &
LPIPS$\downarrow$ &
Latent $\ell_2\downarrow$ &
$\Delta$-LPIPS$\downarrow$ &
Motion err.$\downarrow$ \\
\midrule
Matched MLP conditioning
& 23.402
& 0.258
& 0.113
& 0.482
& 4.429 \\

Wan2.1 initialization
& 24.640
& \textbf{0.235}
& 0.099
& 0.472
& 4.414 \\

\midrule
\textbf{Full model (Ours)}
& \textbf{24.994}
& 0.236
& \textbf{0.095}
& \textbf{0.466}
& \textbf{4.155} \\
\bottomrule
\end{tabular*}
\end{center}
\end{table}

\subsection{UltraWorld for Robotic Ultrasound Planning}
\label{sec:applications}

\begin{figure}[!t]
    \centering

    \begin{minipage}[t]{0.41\linewidth}
        \vspace{0pt}
        \centering

        \vspace{0.55\baselineskip}

        \includegraphics[
            width=\linewidth,
            trim=0 10 0 0,
            clip
        ]{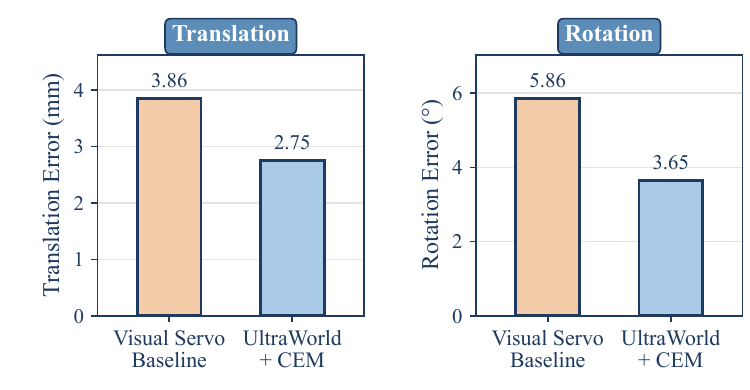}

        \captionof{figure}{Closed-loop planning errors.}
        \label{fig:planning_error}
    \end{minipage}
    \hfill
    \begin{minipage}[t]{0.58\linewidth}
        \vspace{0pt}
        \centering

        \captionof{table}{\textbf{Generalization to real clinical ultrasound.}}
        \label{tab:real_clinical}

        {
        \setlength{\tabcolsep}{2.2pt}
        \renewcommand{\arraystretch}{1.04}

        \begin{tabular*}{\linewidth}{
            @{\extracolsep{\fill}}lcccc@{}
        }
        \toprule

        \multirow{2}{*}{Conditioning} &
        \multicolumn{2}{c}{BUSI} &
        \multicolumn{2}{c}{BUV} \\

        \cmidrule(lr){2-3}
        \cmidrule(l){4-5}

        &
        \shortstack{Zero-\\action$\downarrow$} &
        \shortstack{Round-\\trip$\downarrow$} &
        \shortstack{Zero-\\action$\downarrow$} &
        \shortstack{Round-\\trip$\downarrow$} \\

        \midrule

        Fourier Enc.
        & 0.110
        & 0.231
        & 0.148
        & 0.276 \\

        MLP Enc.
        & 0.172
        & 0.238
        & 0.244
        & 0.253 \\

        PRoPE
        & 0.159
        & 0.227
        & 0.226
        & 0.317 \\

        \textbf{AsMap (Ours)}
        & \textbf{0.042}
        & \textbf{0.062}
        & \textbf{0.059}
        & \textbf{0.076} \\

        \bottomrule
        \end{tabular*}
        }

    \end{minipage}
    \vspace{-0.5\baselineskip}
\end{figure}

\begin{figure}[!t]
\begin{center}
\includegraphics[width=\linewidth]{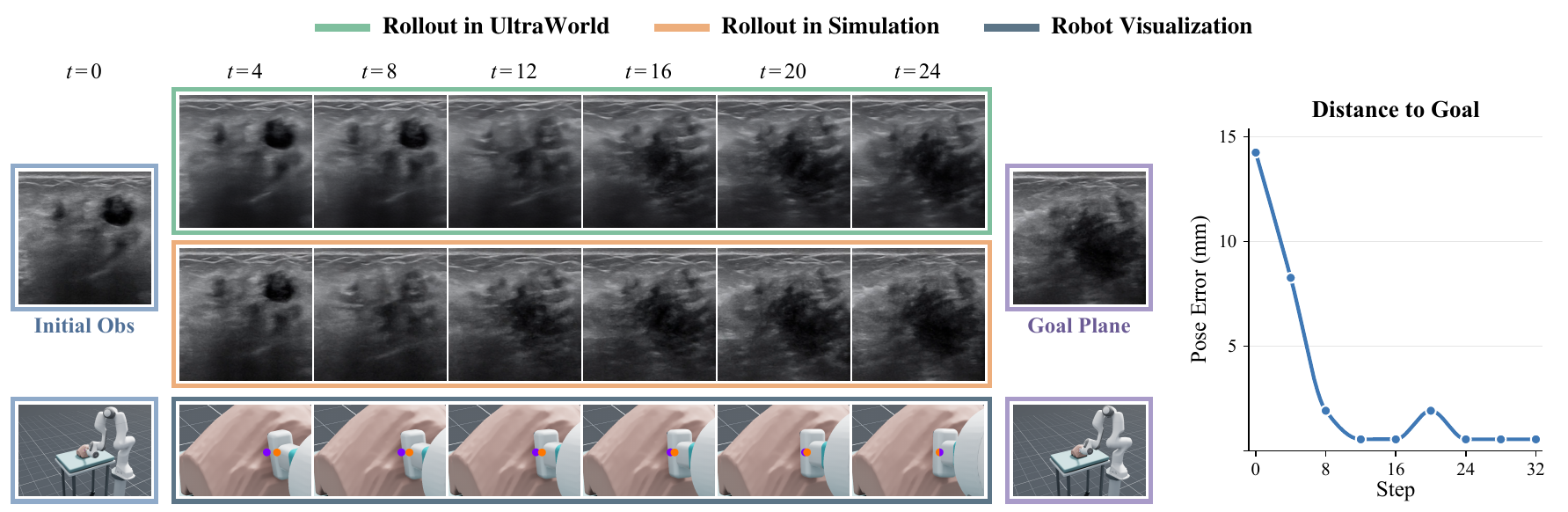}
\end{center}
\caption{
\textbf{Closed-loop robotic ultrasound planning with UltraWorld.}
A representative episode showing imagined rollouts, executed observations,
and the target view. Only the first optimized action is executed before replanning from the new observation.
}
\label{fig:planning_rollout}
\end{figure}

Building on world-model-based robotic planning~\citep{hansen2024tdmpc2,zhu2025irasim,zhou2025dinowm}, we evaluate UltraWorld for goal-directed local ultrasound planning in simulation.
We use model predictive control (MPC)~\citep{mayne2000constrained}, optimizing candidate probe-action sequences with the cross-entropy method (CEM)~\citep{deboer2005tutorial}. We compare against an intensity-based visual-servoing baseline inspired by \citet{nadeau2013intensity}; implementation details are provided in Appendix~\ref{app:planning}.
Across nine episodes, UltraWorld achieves lower mean terminal translation and rotation errors than visual servoing (Fig.~\ref{fig:planning_error}). Figure~\ref{fig:planning_rollout} shows a representative closed-loop episode in which imagined rollouts guide the probe toward the target view.

\subsection{Generalization to Real Clinical Ultrasound}
\label{sec:real_generalization}

We evaluate unseen real initial frames from patient-held-out BUV and
independent BUSIdataset~\citep{aldhabyani2020busi}. Without paired futures, we use zero-action and round-trip
LPIPS to assess stability and recovery consistency~\citep{li2026dworldeval}.
AsMap achieves the lowest errors on both datasets (Table~\ref{tab:real_clinical});
qualitative rollouts are provided in Appendix~\ref{app:results_limitations}.

\section{Conclusion}
\label{sec:conclusion}

We presented UltraWorld, a recipe for learning ultrasound world models from
untracked clinical videos. It combines a mask-conditioned ultrasound video
prior with anatomy-guided trajectory synthesis for self-distillation, and
introduces the Acoustic Sampling Map (AsMap) to represent pixel-wise ultrasound
sampling geometry. Experiments validate the fidelity and controllability of the
video generator, and show that UltraWorld improves prediction fidelity and
acquisition following over alternative conditioning representations. We further
demonstrate closed-loop local scan planning with lower terminal translation and rotation errors than visual servoing, and improved consistency from unseen real clinical observations. Together, these results provide a practical route from untracked clinical videos to action-conditioned ultrasound world models.
Future work will validate the approach with physical
robotic systems.

\clearpage
\subsection*{AI use statement}

In this work, we did not use generative AI tools for tasks requiring mandatory disclosure under the ICLR 2027 AI Policy, including developing theoretical or conceptual frameworks, formulating mathematical claims or proofs, proposing or refining hypotheses, designing research methodology or experiments, implementing the proposed methods, generating synthetic datasets, cleaning or reformatting datasets, conducting qualitative or thematic data analysis, or interpreting experimental results. Translation and proof-related assistance were not applicable to this work.

We used generative AI tools for tasks for which disclosure is recommended, including improving the readability and clarity of the manuscript, assisting with software code editing and debugging, brainstorming, and supporting literature search and summarization.

All AI-assisted text was reviewed and revised by the authors. AI-assisted code was manually inspected and validated through testing and experimental execution. Literature-related suggestions and summaries were checked against the original sources, and all references and factual claims included in the paper were independently verified by the authors. Generative AI outputs were not treated as authoritative sources.

We take responsibility for the final content of this work, including all text, claims, code, results, and artifacts produced with the aid of generative AI.

\subsection*{Ethics statement}

This work uses publicly available datasets and does not involve human subjects or personally identifiable information. We are not aware of any specific ethical concerns beyond those generally associated with the use of machine learning systems.

\subsection*{Reproducibility statement}

We provide detailed descriptions of the proposed method, model architecture, training procedure, hyperparameter settings, dataset preprocessing, and evaluation protocols in the main paper and appendices. Additional implementation details and experimental results are included in the supplementary material. The code and configuration files necessary to reproduce the main experiments will be released upon publication.

\bibliography{iclr2027_conference}
\bibliographystyle{iclr2027_conference}

\clearpage
\appendix

\section{Properties of the Acoustic Sampling Map}
\label{app:asmap_properties}

We provide additional analysis of the Acoustic Sampling Map (AsMap)
introduced in Sec.~\ref{sec:asmap}. AsMap converts the acquisition state
$\xi_t=(T_t,\kappa_t)$ into pixel-wise physical sampling geometry and
expresses the resulting geometry in the coordinate system of the first
observed frame. Below, we clarify several properties of this representation.

\paragraph{Sensitivity to imaging configuration.}
A probe pose alone does not uniquely determine an ultrasound observation,
because the physical samples represented by the image also depend on the
acquisition configuration $\kappa_t$. Consider two acquisition states
\begin{equation}
    \xi_t=(T_t,\kappa_t),
    \qquad
    \xi'_t=(T_t,\kappa'_t),
\end{equation}
that share the same probe pose but differ in their imaging configurations.
Their corresponding AsMaps are
\begin{equation}
    \mathbf{G}_t
    =
    \operatorname{AsMap}(T_t,\kappa_t;T_1),
    \qquad
    \mathbf{G}'_t
    =
    \operatorname{AsMap}(T_t,\kappa'_t;T_1).
\end{equation}
If $\kappa_t$ and $\kappa'_t$ induce different physical sampling locations,
beam directions, or sampling depths, then in general
\begin{equation}
    \mathbf{G}_t \neq \mathbf{G}'_t.
\end{equation}
For example, changing the imaging depth or field of view changes the physical
locations associated with image pixels even if $T_t$ remains fixed.
AsMap therefore distinguishes acquisition changes that are not observable
from a 6-DoF probe pose alone.

\paragraph{Invariance to the global patient coordinate frame.}
AsMap expresses acquisition geometry relative to the probe coordinate system
of the first frame. This construction removes dependence on the particular
global coordinate system used to describe the trajectory.

Consider an arbitrary rigid transformation $S\in SE(3)$ applied to the
patient coordinate system. The probe poses then become
\begin{equation}
    T'_t = S T_t,
    \qquad
    T'_1 = S T_1.
\end{equation}
The relative transformation used to construct AsMap satisfies
\begin{equation}
    (T'_1)^{-1}T'_t
    =
    (S T_1)^{-1}(S T_t)
    =
    T_1^{-1}S^{-1}S T_t
    =
    T_1^{-1}T_t.
    \label{eq:asmap_global_frame_invariance}
\end{equation}
Therefore, the first-frame-relative physical sampling positions and acoustic
beam directions are unchanged under a rigid change of the global patient
coordinate system. Consequently,
\begin{equation}
    \mathbf{G}'_t = \mathbf{G}_t.
\end{equation}
This property is useful because the conditioning describes how the current
acquisition differs geometrically from the observed initial acquisition,
rather than encoding an arbitrary external reference frame.

\paragraph{Independence from numerical pose parameterization.}
AsMap is constructed after the probe pose has been converted to a rigid
transformation $T_t\in SE(3)$. Let $\boldsymbol{\eta}_t$ and
$\boldsymbol{\eta}'_t$ denote two numerical pose representations, and let
$f(\cdot)$ map either representation to $SE(3)$. If
\begin{equation}
    f(\boldsymbol{\eta}_t)
    =
    f(\boldsymbol{\eta}'_t)
    =
    T_t,
\end{equation}
then both representations produce the same relative transformation
$T_1^{-1}T_t$ and therefore the same AsMap.

Thus, after conversion to the physical acquisition geometry, the conditioning
does not depend on whether the same rigid pose was originally represented,
for example, by Euler angles, quaternions, or another equivalent
parameterization. This property concerns the numerical representation of an
identical physical pose and should not be interpreted as invariance to
different probe poses or imaging configurations.

\paragraph{Metric interpretation.}
When the acquisition system is calibrated in physical units,
$\mathbf{q}_{t,u,v}$ represents a physical sampling location,
$\rho_{t,u,v}$ represents a physical sampling depth, and
$\mathbf{b}_{t,u,v}$ is a unit acoustic beam direction. The rigid
first-frame transformation preserves these physical quantities. Therefore,
AsMap retains the physical magnitude of probe translations and changes in
sampling scale rather than representing them only through an arbitrary
normalized action vector.

This metric interpretation assumes metric calibration of both the probe pose
and the mapping from image coordinates to physical sampling geometry.
Without such calibration, AsMap still defines a spatially aligned acquisition
representation, but its position and depth channels should not be interpreted
as absolute physical distances.

\paragraph{Relation to camera-ray conditioning.}
Camera-conditioned video models commonly associate image pixels with optical
viewing rays or encode relative projective geometry. In these representations,
the ray primarily specifies how a pixel observes the scene from a given
camera configuration. Ultrasound acquisition differs because a B-mode pixel
corresponds to a depth-resolved acoustic sample rather than only to a viewing
direction.

AsMap makes this distinction explicit by representing
\begin{equation}
    \left(
        \mathbf{q}_{t,u,v},
        \mathbf{b}_{t,u,v},
        \rho_{t,u,v}
    \right)
\end{equation}
for every image location. The beam direction
$\mathbf{b}_{t,u,v}$ describes the local acoustic propagation direction,
while $\mathbf{q}_{t,u,v}$ and $\rho_{t,u,v}$ specify the corresponding
depth-resolved physical sample. This allows changes in probe pose and imaging
configuration to be converted into a common pixel-aligned sampling
representation before being provided to the video model.

\paragraph{Spatial alignment with visual tokens.}
Global pose encodings associate one acquisition vector with an entire frame.
In contrast, AsMap retains the correspondence between acquisition geometry
and image location. After spatial and temporal alignment by the AsMap encoder
$\Phi$, each geometry feature is associated with the video tokens representing
the corresponding region and acquisition time.

This distinction is especially relevant when an acquisition change induces
spatially non-uniform changes in the observed anatomy. Rather than requiring
the video backbone to infer the mapping from a global pose vector to
pixel-level image changes, AsMap exposes this correspondence explicitly in
the conditioning signal.

\paragraph{Scope of the representation.}
AsMap represents ultrasound \emph{sampling geometry}. In particular, the
representation does not explicitly encode tissue-dependent scattering,
attenuation, speckle formation, or other acoustic interactions responsible
for ultrasound appearance. In UltraWorld, these visual and temporal
characteristics are learned from clinical ultrasound videos by the generative
prior, while AsMap specifies how the prescribed acquisition trajectory changes
the spatial sampling of the underlying anatomy.

This separation is central to our formulation: the learned video prior
provides ultrasound appearance and temporal dynamics, whereas AsMap provides
the acquisition geometry required to transfer these priors to
action-conditioned prediction.

\section{UltraWorld Implementation Details}
\label{app:implementation}

Algorithms~\ref{alg:ultraworld_training}
and~\ref{alg:ultraworld_inference} summarize the training recipe and
action-conditioned inference of UltraWorld.
We denote the video transformer parameters by $\theta$ and the AsMap
encoder parameters by $\psi$.
The pretrained VAE and image/text encoders remain fixed.
For brevity, $\mathcal{C}_1$ groups the inherited first-frame latent,
image features, and text conditioning.
The AsMap encoder $\Phi_\psi$ aligns sampling maps with video tokens,
and its output is added once before the first transformer block.

\begin{algorithm}[t]
\caption{UltraWorld Training Recipe}
\label{alg:ultraworld_training}
\begin{algorithmic}[1]
\Require Pretrained video generator $p_{\phi_0}$;
         untracked clinical videos $\mathcal{D}_{\mathrm{US}}$;
         segmentation model $\mathcal{S}$;
         semantic anatomical volumes $\mathbb{V}$
\Ensure World-model parameters $\theta$ and AsMap encoder parameters $\psi$

\Statex \textbf{1. Learn an ultrasound video prior}
\State Construct video--mask pairs
       $(\mathbf{X}_{1:T},\mathcal{S}(\mathbf{X}_{1:T}))$
       from $\mathcal{D}_{\mathrm{US}}$
\State Associate each pair with a reference ultrasound image $\mathbf{r}$
\State Adapt $p_{\phi_0}$ using LoRA and conditional flow matching
       to obtain $p_{\phi^\star}(\mathbf{X}\mid\mathbf{r},\mathbf{M})$

\Statex \textbf{2. Synthesize acquisition-supervised trajectories}
\State $\widehat{\mathcal{D}}_{\mathrm{traj}}\gets\emptyset$
\For{each trajectory to synthesize}
    \State Sample a volume $\mathcal{V}\in\mathbb{V}$ and a programmed
           acquisition trajectory
           $\boldsymbol{\Xi}_{1:T}
           =\{(\mathbf{T}_t,\boldsymbol{\kappa}_t)\}_{t=1}^{T}$
    \State Obtain anatomical masks by volumetric resampling:
           $\mathbf{m}_t\gets
           \Pi(\mathcal{V};\mathbf{T}_t,\boldsymbol{\kappa}_t)$
    \State Select a clinical reference image $\mathbf{r}$
    \State Generate
           $\widehat{\mathbf{X}}_{1:T}
           \sim p_{\phi^\star}
           (\mathbf{X}_{1:T}\mid\mathbf{r},\mathbf{M}_{1:T})$
    \State Store $(\widehat{\mathbf{X}}_{1:T},
           \boldsymbol{\Xi}_{1:T})$ and associated conditioning metadata
           in $\widehat{\mathcal{D}}_{\mathrm{traj}}$
\EndFor

\Statex \textbf{3. Self-distill into an action-conditioned world model}
\State Initialize the shared video-model weights $\theta$ from $\phi^\star$
\State Initialize $\Phi_\psi$ with a zero-output projection
\For{training stage $s\in\{1,2\}$}
    \State Set trainable parameters:
           $\Omega_1=\{\psi\}$ and $\Omega_2=\{\theta,\psi\}$
    \For{each optimization step}
        \State Sample a synthesized video, its acquisition trajectory,
               and conditioning metadata
        \State Encode the clean video latent
               $\mathbf{z}_0\gets\mathcal{E}
               (\widehat{\mathbf{X}}_{1:T})$
        \State Construct $\mathcal{C}_1$ from the initial frame and
               associated text condition
        \State Build first-frame-relative sampling maps:
               $\mathbf{G}_{1:T}\gets
               \operatorname{AsMap}
               (\boldsymbol{\Xi}_{1:T};\mathbf{T}_1)$
        \State Encode token-aligned geometry features:
               $\mathbf{E}\gets\Phi_\psi(\mathbf{G}_{1:T})$
        \State Randomly drop $\mathbf{E}$ according to the
               stage-specific conditioning-dropout probability
        \State Sample a shifted flow time $\tau$ and
               $\boldsymbol{\epsilon}\sim\mathcal{N}(\mathbf{0},\mathbf{I})$
        \State $\mathbf{z}_\tau\gets
               (1-\tau)\mathbf{z}_0+\tau\boldsymbol{\epsilon}$
        \State Predict
               $\widehat{\boldsymbol{v}}\gets
               \boldsymbol{v}_\theta
               (\mathbf{z}_\tau,\tau\mid\mathcal{C}_1,\mathbf{E})$
        \State $\mathcal{L}\gets
               w(\tau)\,
               \operatorname{MSE}_{\mathrm{future}}
               \bigl(\widehat{\boldsymbol{v}},
                     \boldsymbol{\epsilon}-\mathbf{z}_0\bigr)$
        \State Update $\Omega_s$ with AdamW using $\mathcal{L}$
    \EndFor
\EndFor
\State \Return $(\theta,\psi)$
\end{algorithmic}
\end{algorithm}

Anatomical masks are used only to train the video prior and synthesize
the supervision videos.
During self-distillation, the mask-control channels remain zero,
while the initial-frame conditioning is retained.
$\operatorname{MSE}_{\mathrm{future}}$ denotes the mean squared error
over future latent frames, excluding the observed initial latent frame.

\begin{algorithm}[t]
\caption{Action-Conditioned Inference with UltraWorld}
\label{alg:ultraworld_inference}
\begin{algorithmic}[1]
\Require Initial ultrasound frame $\mathbf{x}_1$;
         prescribed acquisition trajectory $\boldsymbol{\Xi}_{1:T}$;
         text condition $t$; trained parameters $(\theta,\psi)$;
         descending flow schedule $1=\tau_S>\cdots>\tau_0=0$
\Ensure Predicted future ultrasound frames
        $\widehat{\mathbf{X}}_{2:T}$

\State Construct initial-frame and text conditioning
       $\mathcal{C}_1\gets
       \operatorname{I2VCondition}(\mathbf{x}_1,t)$
\State $\mathbf{G}_{1:T}\gets
       \operatorname{AsMap}(\boldsymbol{\Xi}_{1:T};\mathbf{T}_1)$
\State $\mathbf{E}\gets\Phi_\psi(\mathbf{G}_{1:T})$
\State Initialize the complete video latent
       $\mathbf{z}_{\tau_S}\sim\mathcal{N}(\mathbf{0},\mathbf{I})$
\For{$i=S,\ldots,1$}
    \State $\boldsymbol{v}\gets
           \boldsymbol{v}_\theta
           (\mathbf{z}_{\tau_i},\tau_i\mid\mathcal{C}_1,\mathbf{E})$
    \State $\mathbf{z}_{\tau_{i-1}}\gets
           \mathbf{z}_{\tau_i}
           +(\tau_{i-1}-\tau_i)\boldsymbol{v}$
\EndFor
\State Decode
       $\widehat{\mathbf{X}}_{1:T}\gets
       \mathcal{D}_{\mathrm{VAE}}(\mathbf{z}_{\tau_0})$
\State \Return $\widehat{\mathbf{X}}_{2:T}$
\end{algorithmic}
\end{algorithm}

\section{Robotic Ultrasound Planning Details}
\label{app:planning}

Algorithm~\ref{alg:ultraworld_planning} summarizes our closed-loop
planning implementation.
The planner interacts with an ultrasound scanning environment that provides
the current observation and supports feasible probe motion.
At each planning step, the planner accesses the current observation, the
goal image, and the motion information required for planning, but not the
target pose or future observations.
UltraWorld remains frozen throughout planning.

\begin{algorithm}[t]
\caption{Closed-Loop Ultrasound Planning with UltraWorld}
\label{alg:ultraworld_planning}
\begin{algorithmic}[1]
\Require Frozen UltraWorld $\mathcal{W}$; scan environment
         $\mathcal{E}$; initial state $i$; goal image $\mathbf{x}_g$
\Ensure Executed observation--action sequence $\mathcal{R}$

\State $\mathbf{x}\gets\operatorname{Observe}(\mathcal{E},i)$;
       $\mathcal{R}\gets\emptyset$

\For{MPC iteration $t=1,\ldots,8$}
    \State Initialize four categorical action distributions uniformly
           over five motion levels
    \State Construct feasible four-block action sequences under the
           environment motion constraints
    \State $\mathcal{B}\gets\emptyset$
           \Comment{Evaluated candidates}

    \For{CEM iteration $\ell=1,\ldots,3$}
        \State Sample up to 16 unevaluated feasible plans using the
               current action distributions
        \State Include the stationary plan in the first iteration

        \For{each sampled plan $\mathbf{U}$}
            \State Repeat each block action for four steps to obtain
                   a 16-step motion trajectory
            \State Construct AsMap conditions $\mathbf{G}^{\mathbf{U}}$
                   relative to the current probe pose
            \State Predict
                   $\widehat{\mathbf{X}}^{\mathbf{U}}_{0:16}
                   \gets\mathcal{W}
                   (\mathbf{x},\mathbf{G}^{\mathbf{U}})$
                   using a shared sampling seed
            \State $J(\mathbf{U})\gets
                   0.7\,d_{\mathrm{LPIPS}}
                   (\hat{\mathbf{x}}^{\mathbf{U}}_4,\mathbf{x}_g)
                   +0.3\,d_{\mathrm{LPIPS}}
                   (\hat{\mathbf{x}}^{\mathbf{U}}_8,\mathbf{x}_g)$
            \State Add the plan and its cost to $\mathcal{B}$
        \EndFor

        \State Select the four lowest-cost plans from all candidates
               in $\mathcal{B}$
        \State Update action distributions using smoothed elite
               frequencies and uniform exploration
    \EndFor

    \State Select the lowest-cost evaluated plan $\mathbf{U}^{\star}$
           from $\mathcal{B}$
    \State Execute only its first four-step block in $\mathcal{E}$
    \State Update $i$ to the reached state and obtain
           $\mathbf{x}\gets\operatorname{Observe}(\mathcal{E},i)$
    \State Append the executed actions and observations to $\mathcal{R}$
\EndFor

\State \Return $\mathcal{R}$
\end{algorithmic}
\end{algorithm}

Each candidate is predicted jointly from the current observation.
Only the selected prefix is executed, after which the planner obtains a new
observation from the environment and replans.
The procedure runs for a fixed budget of eight MPC iterations without
early termination or additional MR resampling.

\section{Additional Experimental Details}
\label{app:experiments}

\subsection{Data Preparation and Evaluation Sets}
\label{app:data}

\paragraph{Data sources and splits.}
BUV~\citep{lin2022buv} and CAMUS~\citep{leclerc2019camus} provide
untracked clinical ultrasound videos for learning the mask-conditioned
ultrasound video prior, covering breast ultrasound and echocardiography,
respectively. The videos are paired with frame-wise anatomical masks.
I-SPY1~\citep{newitt2016ispy1,hylton2016ispy1} provides volumetric breast
DCE-MRI and lesion annotations used to construct anatomy-guided
acquisition trajectories. All datasets are split at the patient level, and
derived clips remain in the partition of their source patient.

\paragraph{Synthetic trajectory preparation.}
Programmed acquisition trajectories determine the anatomical cross-sections
resampled from each I-SPY1 volume. The resulting mask trajectories are rendered
into ultrasound videos by the mask-conditioned generator, yielding paired
video--acquisition trajectories for world-model training and evaluation.
World-model clips contain 17 frames at $512\times512$ resolution.
Forward and backward subsequences use temporal strides of one or two and are
supplemented with palindromic and stationary trajectories.
RGB intensities are normalized to $[-1,1]$ before VAE encoding.
Video latents, first-frame image features, text embeddings, and synchronized
acquisition states are precomputed for training.

\paragraph{Real-clinical evaluation sets.}
Real-initial-frame experiments use the patient-held-out BUV split and the
independent BUSI dataset. Each clinical image serves as the initial observation,
and the world model predicts a subsequent sequence under a prescribed
acquisition trajectory. No paired clinical future observations are available,
so these experiments evaluate observation stability and recovery consistency
rather than direct prediction against a ground-truth future.

\subsection{Baseline and Ablation Settings}
\label{app:baselines}

\paragraph{Action-conditioning baselines.}
All conditioning variants use the same video backbone and receive the same
acquisition state, initial-frame observation condition, and text condition.
Fourier Enc.~\citep{fan2026actionconditioned} and
MLP Enc.~\citep{yue2025echoworld} convert frame-level acquisition
information into global embeddings that are broadcast across spatial tokens
and added to the input video tokens.
AsMap instead produces spatially aligned token features.
PRoPE~\citep{li2025prope} applies projective geometric conditioning inside
self-attention and is therefore not converted to the same additive
token-conditioning interface. Table~\ref{tab:encoder_architectures} summarizes the action encoders used in our comparisons. Parameter counts include only the additional
conditioning encoder and exclude the shared video backbone.

\begin{table}[t]
\caption{
\textbf{Architectures of token-level action encoders.}
}
\label{tab:encoder_architectures}
\begin{center}
\resizebox{\linewidth}{!}{
\begin{tabular}{llr}
\hline
\textbf{Encoder} &
\textbf{Architecture} &
\textbf{Parameters} \\
\hline

Fourier Enc.
& Fourier features + MLP
& 1.11M \\

MLP Enc.
& MLP
& 0.79M \\

MLP Enc. (Matched)
& Matched-size MLP
& 53.49M \\

PRoPE
& 30-layer Linear
& 70.82M \\

\textbf{AsMap (Ours)}
& Convolution + ResBlock
& 53.48M \\
\hline
\end{tabular}
}
\end{center}
\end{table}

\paragraph{Controlled ablations.}
The matched MLP uses the same first-frame-relative coordinate convention as
AsMap and approximately matches its encoder parameter count. This ablation therefore controls for encoder capacity and reference-frame
convention without imposing an identical temporal representation.

For the initialization ablation, we retain AsMap and exactly the same
synthesized action--video trajectories while replacing initialization from the
mask-conditioned ultrasound generator with initialization from the
corresponding Wan2.1 base checkpoint.
This isolates the effect of prior-weight inheritance while keeping the
generator-produced supervision fixed.

\paragraph{Training and inference configuration.}
\label{app:training_details}
The mask-conditioned ultrasound generator is initialized from
Wan2.1-FunControl-1.3B-V1.1~\citep{wan2025}.
All models are trained on a single NVIDIA H200 GPU using
$512\times512$ videos with 17 frames per clip and a batch size of 8.
We use AdamW~\citep{loshchilov2019decoupled} for optimization.

The mask-conditioned generator is trained for 2K steps with a learning rate
of $1\times10^{-4}$.
The ultrasound world model is trained for 10K steps with a learning rate of
$2\times10^{-6}$.
As described in Sec.~4.3, world-model adaptation first trains the acquisition
encoder with the video backbone frozen and then jointly fine-tunes the
acquisition encoder and video backbone.

At inference, all methods use 25 Euler sampling steps with a timestep-shift
factor of 5. Text and acquisition guidance scales are both set to 1.

\subsection{Evaluation Metrics and Protocols}
\label{app:metrics}

\paragraph{Prediction fidelity.}
For world-model fidelity evaluation, the conditioning frame is excluded from
all reported metrics, and videos decoded from clean latents are used as
references. We report PSNR, LPIPS, and latent $\ell_2$ distance.
LPIPS uses an AlexNet backbone, and latent $\ell_2$ is implemented as
mean-squared error in latent space.

\paragraph{Segmentation-based generator evaluation.}
We apply a frozen UltraSam segmenter~\citep{meyer2026ultrasam} to generated and reference ultrasound
frames using identical box prompts derived from connected lesion components in
the conditioning masks.
Mean IoU between segmentations of generated and reference frames measures
anatomical consistency, while mean Dice between generated segmentations and
conditioning masks measures mask controllability.
Frames with no lesion in the conditioning mask are excluded.
No empty predictions occurred in this evaluation.

\paragraph{Action following.}
Following~\citet{li2026dworldeval}, we compute $\Delta$-LPIPS over a
four-frame interval. For reference and predicted videos, respectively,
\begin{equation}
    \mathbf{D}_t
    =
    \mathbf{x}_{t+4}-\mathbf{x}_t,
    \qquad
    \widehat{\mathbf{D}}_t
    =
    \hat{\mathbf{x}}_{t+4}-\hat{\mathbf{x}}_t.
\end{equation}
Each difference image is independently RMS-normalized:
\begin{equation}
    \Delta\text{-LPIPS}
    =
    \frac{1}{|\mathcal{P}_4|}
    \sum_{t\in\mathcal{P}_4}
    d_{\mathrm{LPIPS}}
    \left(
        \mathcal{N}(\widehat{\mathbf{D}}_t),
        \mathcal{N}(\mathbf{D}_t)
    \right),
    \qquad
    \mathcal{N}(\mathbf{D})
    =
    \frac{\mathbf{D}}
    {\sqrt{\operatorname{mean}(\mathbf{D}^{2})+\epsilon}},
\end{equation}
where $\epsilon=10^{-8}$.
The set $\mathcal{P}_4$ contains valid four-frame pairs whose reference
difference exceeds the static RMS threshold.
Scores are first averaged over valid pairs within each video and then across
videos; videos with no valid pairs are excluded.

\emph{Motion Error}, defined in Eq.~\ref{eql:motion_error}, compares
adjacent-frame target-anatomy displacements.
Reference and predicted lesion centroids are obtained using UltraSAM with
identical annotation-derived box prompts. The annotation masks are used only
to construct prompts and do not provide the reference centroid locations.
The set $\mathcal{P}$ in Eq.~\ref{eql:motion_error} contains adjacent-frame
pairs for which both reference and predicted centroids exist at both time
points. Motion errors are averaged over valid pairs within each video and then
over valid videos.

\paragraph{Real-clinical consistency.}
Real-initial-frame experiments are evaluated separately on BUSI and BUV.
Zero-action LPIPS measures the average perceptual distance between the real
input observation and predicted future frames while the acquisition remains
stationary.

Round-trip LPIPS uses a 17-frame single-axis forward-and-return trajectory,
$0\rightarrow8\rightarrow0$, and compares the real input observation with the
final predicted frame after returning to the starting acquisition state.
Both metrics use the real input frame, rather than a generated conditioning
frame, as the reference.
Results are averaged over cases within each dataset.
These metrics evaluate observation stability and return consistency without
requiring paired clinical futures; round-trip consistency alone does not
establish action-following accuracy.

\paragraph{Robotic planning evaluation.}
We report terminal translation error and terminal rotation error.
Let $(\mathbf{p}_f,\mathbf{R}_f)$ and
$(\mathbf{p}_g,\mathbf{R}_g)$ denote the final executed and target probe
poses. The terminal errors are
\begin{equation}
\begin{aligned}
    e_{\mathrm{pos}}
    &=
    \|\mathbf{p}_f-\mathbf{p}_g\|_2,\\
    e_{\mathrm{rot}}
    &=
    \frac{180}{\pi}
    \left\|
        \operatorname{rotvec}
        (\mathbf{R}_g^\top\mathbf{R}_f)
    \right\|_2 ,
\end{aligned}
\label{eq:planning_errors}
\end{equation}
reported in millimeters and degrees, respectively.
Both metrics are evaluated at the end of the fixed execution budget.
The target pose is used only for evaluation; candidate selection during
planning uses predicted-image LPIPS and never accesses the target pose.

\section{Additional Results and Limitations}
\label{app:results_limitations}

\subsection{Mask-Conditioned Generator Evaluation}
\label{app:generator_eval}

\paragraph{Generation fidelity.}
We evaluate the mask-conditioned ultrasound generator using both image-level
and task-aware metrics.
Generated videos achieve an SSIM of 0.4265 and an LPIPS of 0.4071.
Following prior work evaluating synthetic ultrasound through downstream
analysis~\citep{yu2026busgen,stojanovski2023echo}, we additionally apply a
pretrained ultrasound segmenter to generated and ground-truth frames.
The resulting segmentations achieve a mean IoU of 0.8736, indicating that the
generated videos preserve the relevant anatomical structures.

\paragraph{Mask controllability.}
We further evaluate whether the generated anatomy follows the prescribed mask
trajectories.
Comparing lesion regions segmented from generated frames with the conditioning
masks yields a Dice score of 0.9260, compared with 0.9538 between
ground-truth segmentations and the same masks.
Frames without lesions in the conditioning masks are excluded.
Metric and segmentation protocols are detailed in
Appendix~\ref{app:metrics}.

\subsection{Action-Conditioned Appearance Dynamics}
\label{app:appearance_dynamics}

A local 2D ultrasound observation can be compatible with different underlying 3D anatomies, leaving future sections only partially determined. Our focus is therefore on learning how scanning operations change ultrasound and anatomical appearance. Synthetic action--video pairs provides explicit supervision for these changes, while self-distillation transfers the resulting action--observation relationships into a model that predicts from local observations and actions.

\begin{figure}[t]
\begin{center}
\includegraphics[width=\linewidth]{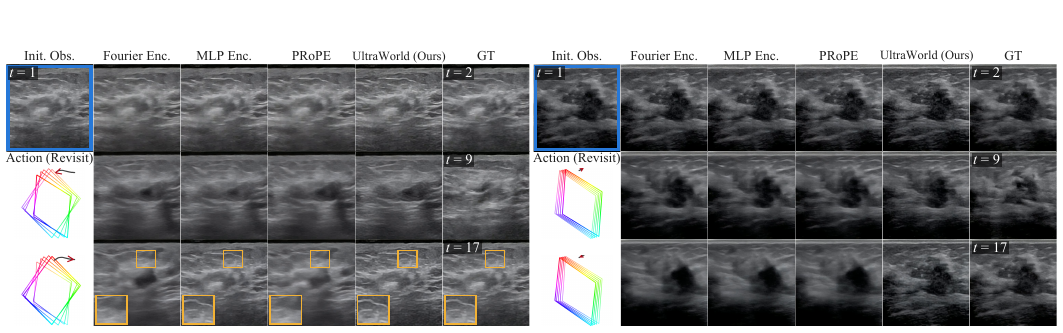}
\end{center}
\caption{\textbf{Qualitative comparison under revisiting.} UltraWorld and alternative conditioning methods predict future ultrasound frames from the same initial observation and prescribed trajectory. Comparisons with reference futures highlight differences in action following and the continuity of local tissue appearance.}
\label{fig:wm_quality_additional}
\end{figure}

Figure~\ref{fig:wm_quality_additional} extends Figure~\ref{fig:wm_quality} to revisiting. UltraWorld shows more consistent responses to the prescribed operations while retaining plausible hypoechoic regions and surrounding tissue appearance. These results illustrate the transfer from generation under explicit anatomical control to predicting motion-induced visual changes while maintaining coherent local structure.

\subsection{Additional world-model rollouts for robotic planning}
\label{app:additioned rollouts}

\begin{figure}[!t]
\begin{center}
\includegraphics[width=\linewidth]{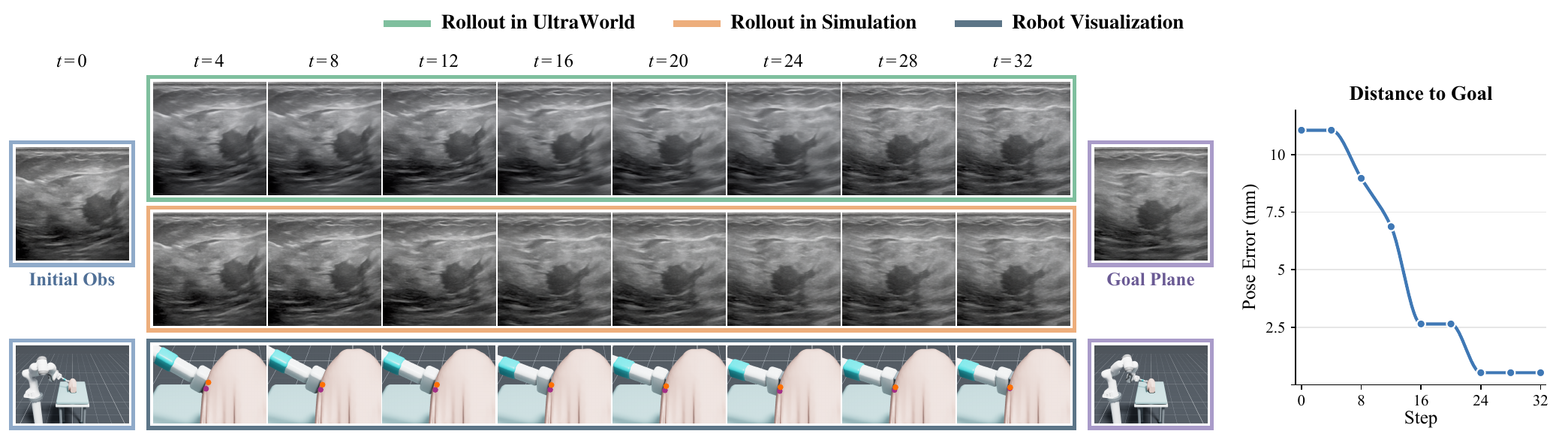}
\end{center}
\caption{
\textbf{Additional qualitative rollouts for robotic planning.}
Additional closed-loop examples showing UltraWorld rollouts and the
corresponding simulator executions, together with the probe-pose error to the
goal during planning.}
\label{fig:additional_rollouts}
\end{figure}

Figure~\ref{fig:additional_rollouts} provides additional closed-loop robotic
planning examples, showing UltraWorld rollouts together with the corresponding
simulator executions and probe-pose errors to the goal over the planning
process. These examples complement the representative episode shown in
Fig.~\ref{fig:planning_rollout} of the main paper.

\subsection{Qualitative Generalization to Real Clinical Ultrasound}

\begin{figure}[!t]
\begin{center}
\includegraphics[width=\linewidth]{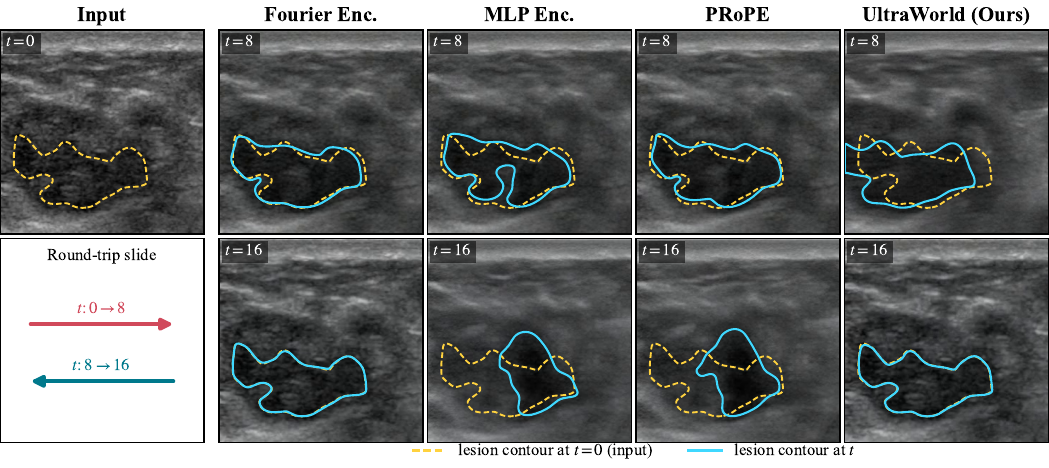}
\end{center}
\caption{
\textbf{Action-conditioned rollouts from real clinical ultrasound observations.}
Each method is initialized from the same real ultrasound image and follows a
prescribed acquisition trajectory.
The intermediate predictions visualize how the predicted anatomy evolves in
response to the prescribed action.
}
\label{fig:real_generalization}
\end{figure}

To complement the quantitative real-clinical evaluation in
Table~\ref{tab:real_clinical}, we visualize action-conditioned predictions
initialized from unseen clinical ultrasound images in
Fig.~\ref{fig:real_generalization}.
The qualitative examples illustrate how the predicted anatomy evolves in
response to prescribed acquisition actions, while the round-trip results in
Table~\ref{tab:real_clinical} evaluate whether the prediction returns close to
the initial observation after a forward-and-return trajectory.
Because paired future observations are unavailable, these experiments assess
action-sensitive behavior and recovery consistency rather than direct
future-prediction fidelity.

\subsection{Counterfactual Acquisition Trajectories}

\begin{figure}[!t]
\begin{center}
\includegraphics[width=\linewidth]{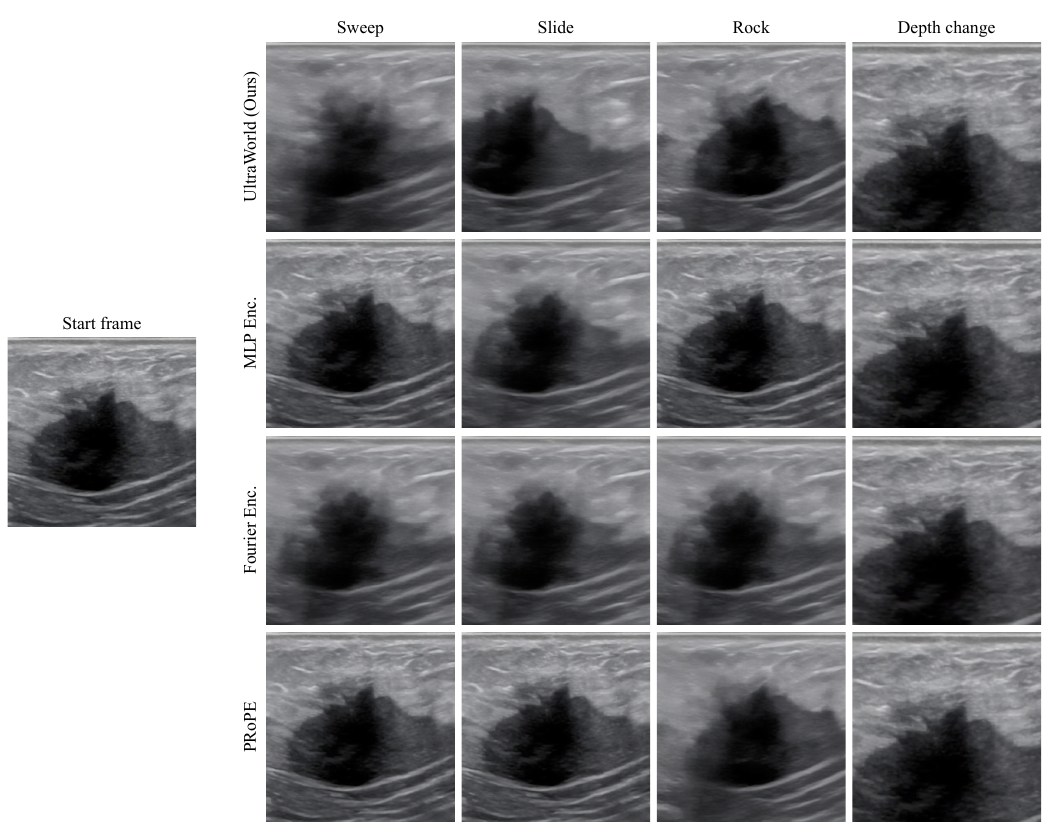}
\end{center}
\caption{
\textbf{Counterfactual predictions under alternative acquisition trajectories.}
Starting from the same ultrasound observation, we prescribe sweep, slide,
rock, and depth-change trajectories and compare the resulting predictions
across action conditioning methods.
}
\label{fig:counterfactual}
\end{figure}

We additionally evaluate action sensitivity by applying alternative prescribed
acquisition trajectories to the same initial observation, as illustrated in
Fig.~\ref{fig:counterfactual}. These counterfactual examples isolate the effect
of the conditioning trajectory: the visual context is fixed, while the
requested probe motion or imaging configuration is changed. As shown in
Fig.~\ref{fig:counterfactual}, different prescribed trajectories lead to
different predicted anatomical evolutions, highlighting whether each action
conditioning method can produce distinct and plausible motion-dependent
responses.

\subsection{Limitations.}

UltraWorld still has several limitations. First, the world model is trained primarily with generator-synthesized action--video trajectories rather than real ultrasound sequences with synchronized probe tracking, so errors or biases in the generative prior may propagate to the distilled world model. Second, AsMap represents acquisition geometry but does not explicitly model tissue-dependent acoustic effects such as scattering, attenuation, and speckle formation, and its metric interpretation assumes calibrated probe poses and imaging geometry. Third, evaluation on real clinical ultrasound is limited to stability and round-trip consistency because paired future observations with known actions are unavailable, while closed-loop robotic planning is currently validated only in simulation and on a limited number of episodes. Future work will explore using a small amount of tracked clinical data for calibration and validation, while preserving the scalability of learning from large-scale untracked clinical videos, as well as extending to broader anatomical and device domains and physical robotic systems.

\end{document}